\documentclass[runningheads]{llncs}
\usepackage[T1]{fontenc}
\usepackage{graphicx,verbatim}
\newcommand{\etal}{\textit{et~al.}}
\begin{document}
\title{Emerging Semantic Segmentation from Positive and Negative Coarse Label Learning}

\author{
Le Zhang \inst{5} \thanks{This research was carried out at the University of Oxford, where the author was affiliated during the period in which this work was completed.}
\and Fuping Wu\inst{1,2}
\and Arun Thirunavukarasu\inst{3}
\and Kevin Bronik\inst{4}
\and Thomas Nichols\inst{1}
\and Bart{\l}omiej W. Papie{\.z}\inst{1,2}
}%% Added for anonymized MICCAI 2025 submission
\authorrunning{L. Zhang et al.}
\institute{Big Data Institute, Li Ka Shing Centre for Health Information and Discovery, \\ University of Oxford, Oxford, UK\\
\and 
Nuffield Department of Population Health, University of Oxford, Oxford, UK\\
\and 
Nuffield Department of Clinical Neurosciences, University of Oxford, Oxford, UK\\
\and 
Department of Engineering Science, University of Oxford, Oxford, UK\\
\and 
School of Engineering, College of Engineering and Physical Sciences,\\ University of Birmingham, Birmingham, UK \\
     % \email{l.zhang.16@bham.ac.uk}
}

\maketitle              % typeset the header of the contribution
\begin{abstract}
Large annotated datasets are vital for training segmentation models, but pixel-level labeling is time-consuming, error-prone, and often requires scarce expert annotators, especially in medical imaging. In contrast, coarse annotations are quicker, cheaper, and easier to produce, even by non-experts. In this paper, we propose to use coarse drawings from both positive (target) and negative (background) classes in the image, even with noisy pixels, to train a convolutional neural network (CNN) for semantic segmentation. We present a method for learning the true segmentation label distributions from purely noisy coarse annotations using two coupled CNNs. The separation of the two CNNs is achieved by high fidelity with the characters of the noisy training annotations. We propose to add a complementary label learning that encourages estimating negative label distribution. To illustrate the properties of our method, we first use a toy segmentation dataset based on MNIST. We then present the quantitative results of experiments using publicly available datasets: Cityscapes dataset for multi-class segmentation, and retinal images for medical applications. In all experiments, our method outperforms state-of-the-art methods, particularly in the cases where the ratio of coarse annotations is small compared to the given dense annotations.

% Authors must provide keywords and are not allowed to remove this Keyword section.

\end{abstract}
\section{Introduction}

Thanks to the availability of large datasets with accurate annotations, fully supervised learning (FSL), especially deep supervised learning, has been translated from theoretical algorithms to practice \cite{zhang2025diffuseg} \cite{bronik2024conditional}. 
However, it is generally expensive, time-consuming, and often infeasible to collect pixel-level labels for large-scale datasets. This problem is particularly prominent in the clinical domain where labeled data are scarce due to the high cost of annotations \cite{zhang2020disentangling}. For instance, accurate segmentation of vessels in fundus retinal images is difficult even for experienced experts due to variability of vessel's location, size, and shape across population or disease. The labeling process is prone to errors, almost inevitably leading to noisy datasets as seen in machine learning benchmark datasets \cite{peterson2019human}. Labeling errors can occur due to automated label extraction, ambiguities in input and output spaces, or human errors (e.g.~lack of expertise). As a consequence, despite the availability of large imaging repositories, the generation of the curated labels that are available to machine learning remains a challenging issue, necessitating the development of methods that learn robustly from noisy annotations. 

% \begin{figure}[t]
% \includegraphics[width=\linewidth]{png/introduction.png}
% \caption{Illustration of a segmentation example using the noisy positive and negative coarse annotations.}
% \label{introduction}
% \end{figure}

To reduce the workload of pixel-level annotation, there has been a considerable effort to exploit weakly-supervised strategies. Weakly-supervised learning (WSL) uses annotations that are cheaper to produce than pixel-wise labels such as bounding boxes \cite{dai2015boxsup,Jiang_2022_CVPR}, coarse annotation \cite{wang2021label,saha2022improving}, scribbles \cite{Lin2016ScribbleSupSC,Zhang_2022_CVPR}, or image-level labels \cite{papandreou2015weakly,pathak2014fully} to train the segmentation models. However, the information from weak annotations is of lower precision and usually suffers from noisy information when curating the labels. For example, image-level labels cannot provide the position information of the object of interest (OOI). 
Using bounding boxes helps to indicate the rough positions of the OOI, but the pixels inside the bounding box may belong to multiple classes if the box size is large.
A more annotator-friendly method of efficient supervision is scribble-based annotation, which only requires the annotator to draw a few lines to mark a small part of the OOI. 
Coarse annotations can provide much more information than scribble and avoid large non-target pixels being grabbed into the bounding box. Meanwhile, drawing coarse annotations on images needs only similar effort and time as the scribble and box-level labeling, and can be conducted by non-experts. Therefore, learning from those coarse annotations, and then correcting the noisy pixels with computational methods, may represent an optimally efficient means of enriching labeled large-scale data with minimal effort.

\textbf{Our contribution:} We introduce an end-to-end supervised segmentation method that estimates true segmentation labels from noisy coarse annotations. The proposed architecture (see Fig.~\ref{pipeline}) consists of two coupled CNNs where the first CNN estimates the true segmentation probabilities, and the second CNN models the characteristics of two different coarse annotations by estimating the pixel-wise confusion matrices (CMs) on a per-image basis. 
Unlike previous WSL methods using coarse annotations, our method models and disentangles the complex mappings from the input images to the noisy coarse annotations and to the true segmentation label simultaneously. Specifically, we model the noisy coarse annotation for the objective along with the complementary label learning for the background or non-objective to enable our model to disentangle robustly the errors of the given annotations and the true labels, even when the ratio of coarse annotation is small (e.g., given scribble for each class). In contrast, this would not be possible with the other WSL methods where the parameters of each coarse annotation are estimated on every target image separately.

\section{Method}

\subsection{Problem Set-up}

In this work, we consider developing a supervised segmentation model by learning the positive (e.g.~object to be segmented) and negative (e.g.~object to be not segmented) coarse annotations that are easy and less expensive to be acquired from annotators. Specifically, we consider a scenario where set of images $\{\textbf{x}_n \in \mathbb{R}^{W\times H\times C}\}_{n=1}^N$ (with $W, H, C$ denoting the width, height and channels of the image) are assigned with coarse segmentation labels $\{\{\tilde{\textbf{y}}_n^{(o)}, \tilde{\textbf{y}}_n^{(c)} \} \in \mathcal{Y}^{W\times H}\}_{n=1,...,N}^{\{o,c\}\in S(\mathbf{x}_i)}$ from objective and complementary categories where $\tilde{\textbf{y}}_n^{(o)}$ and $\tilde{\textbf{y}}_n^{(c)}$ denote the noisy objective and complementary coarse annotations, respectively and $S(\mathbf{x}_n)$ denotes the set of all annotations of image $\textbf{x}_i$ and $\mathcal{Y}=[1, 2,...,L]$ denotes the set of segmentation classes.

\begin{figure*}[t]
\centering
\includegraphics[width=0.95\linewidth]{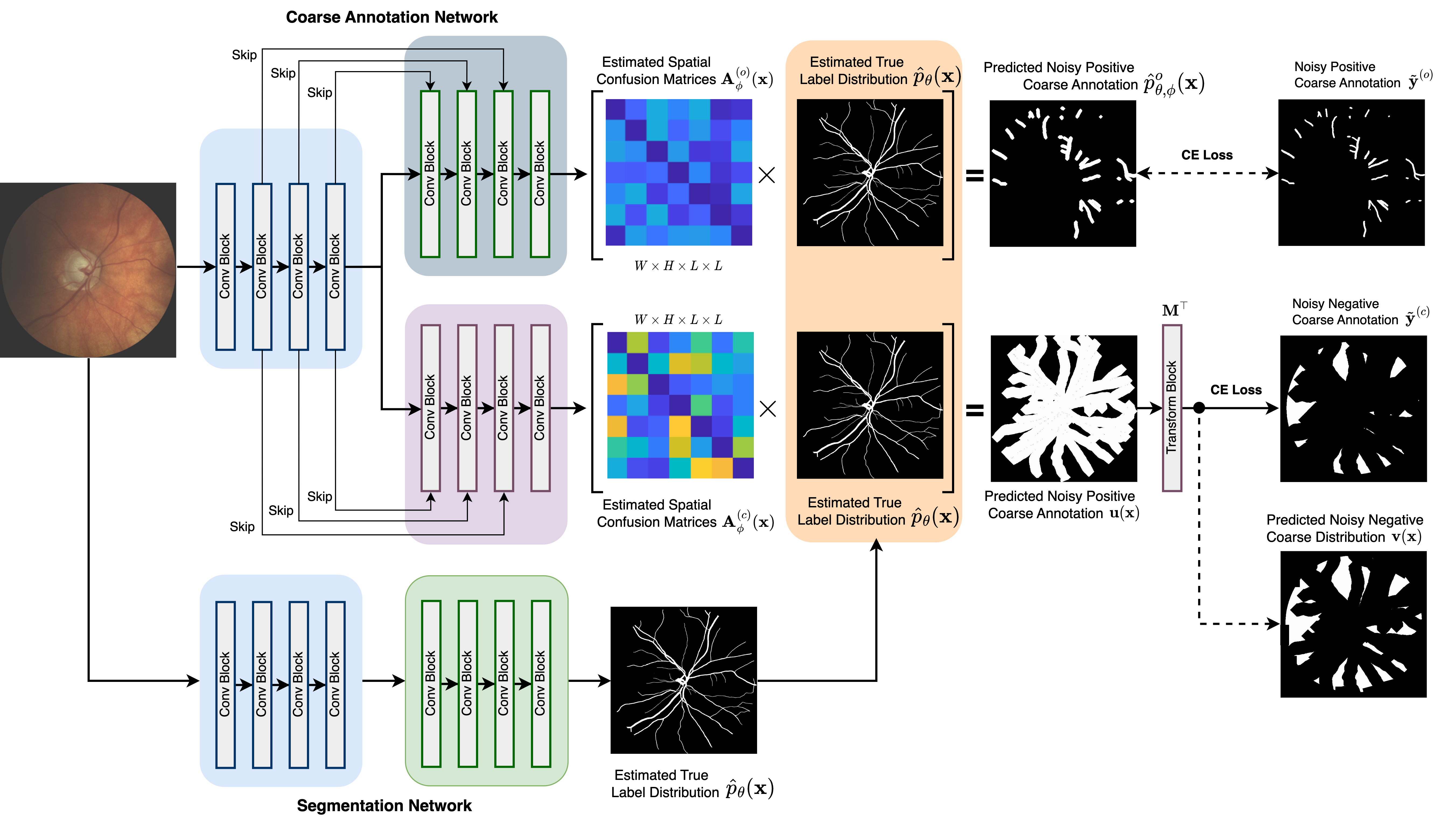}
\caption{General schematic of the model supervised by noisy coarse annotations. The method consists of two components: (1) Segmentation network parameterized by $\theta$ that generates an estimate of the true segmentation probabilities $\hat{p}_\theta (\textbf{x})$ for the given input image $\textbf{x}$; (2) Coarse annotation network, parameterized by $\phi$, that estimates the confusion matrices (CMs) $\{\textbf{A}_{\phi}^{(o)}(\textbf{x}), \hspace{1mm} \textbf{A}_{\phi}^{(c)}(\textbf{x})\}$ of the noisy coarse annotations.}
\label{pipeline}
\end{figure*}

\subsection{Learning from Noisy Coarse Annotations}

In this section, we describe how we jointly optimize the parameters of the segmentation network, $\theta$, and the parameters of the coarse annotation network, $\phi$. In short, we minimize the negative log-likelihood of the probabilistic model for both positive and negative coarse annotations via stochastic gradient descent. A detailed description is provided below. 

\noindent\textbf{Learning with Positive Coarse Label.} Given training input $\textbf{X}=\{\textbf{x}_n\}_{n=1}^N$ and positive coarse labels ${\tilde{\textbf{Y}}^{\{o\}}} = \{ \tilde{ \textbf{y}}_n^{\{o\}}: \{o\} \in S(\textbf{x}_n)\} _{n = 1}^N$, we optimize the parameters $\{ \theta , \phi \}$ by minimizing the negative log-likelihood (NLL), $ - \log p({\tilde{\textbf{Y}}^{(1)}},..., {\tilde{\textbf{Y}}^{(O)}}\left| \textbf{X} \right.)$. The optimization objective equates to the sum of cross-entropy (CE) losses between the observed positive coarse segmentation and the predicted label distributions:
\begin{equation}
\begin{aligned}
    &- \log p({\tilde{\textbf{Y}}^{(1)}},...,{\tilde{\textbf{Y}}^{(O)}}\left| \textbf{X} \right.) \\
    &= \sum\limits_{n = 1}^N {\sum\limits_{o = 1}^O  \mathds{1}(o \in \mathcal{S}({\textbf{x}_n}))}  \cdot
    \text{CE}\big{(}\hat{\textbf{A}}_{\phi}^{(o)}(\textbf{x}_n)\cdot \hat{\textbf{p}}_\theta ({\textbf{x}_n}), \hspace{1mm} \tilde{\textbf{y}}_n^{(o)}\big{)} 
    \label{eq3}
\end{aligned}
\end{equation}

Minimizing Eq.~\eqref{eq3} encourages the positive-specific predictions $\hat{\textbf{p}}_{\theta, \phi}^{(o)}(\textbf{x})$ to be as close as possible to the provided positive coarse label distributions ${\textbf{p}^{(o)}}(\textbf{x})$. 
However, this loss function alone cannot separate the annotation noise from the true label distribution; there are many combinations of pairs $ {\hat{\textbf{A}}_{\phi}}(\mathbf{x})$ and segmentation model $\hat{\textbf{p}}_\theta(\mathbf{x})$ such that $\hat{\textbf{p}}_{\theta, \phi}(\textbf{x})$ matches well the provided distribution $\textbf{p}(\mathbf{x})$ for any input image $\textbf{x}$ (e.g.~permutations of rows in the CMs). 
Tanno \etal \cite{tanno2019learning} addressed an analogous issue for the classification task, and here we add the trace of the estimated CMs to the loss function for positive coarse annotation in Eq.(\ref{eq3}) as a regularisation term. We thus optimize the combined loss:
\begin{equation}
\begin{aligned}
     \mathcal{L}_{\text{obj}}(\theta, \phi) &:= {\mathcal{L}}_{\text{obj}}(\theta, \phi)(\hat{\textbf{A}}_{\phi}^{(o)}(\textbf{x}_n)\cdot \hat{\textbf{p}}_\theta ({\textbf{x}_n}), \tilde{y}^{(o)})  \\ 
    &:= \sum\limits_{n = 1}^N {\sum\limits_{o = 1}^O \mathds{1}(o \in \mathcal{S}({\textbf{x}_n}))} \cdot \Big{[} \text{CE}\big{(}\hat{\textbf{A}}_{\phi}^{(o)}(\textbf{x}_n)\cdot \hat{\textbf{p}}_\theta({\textbf{x}_n}),\hspace{1mm} \tilde{\textbf{y}}_n^{(o)}\big{)}\hspace{1mm} \\
    & + \hspace{1mm}\lambda\cdot \text{tr}\big{(}\hat{\textbf{A}}^{(o)}_{\phi}(\textbf{x}_n)\big{)}\Big{]}
    \label{eq4}
\end{aligned}
\end{equation}
where $\mathcal{S}({\textbf{x}})$ denotes the set of all positive coarse labels available for image $\textbf{x}$, and $\text{tr}({\textbf{A}})$ denotes the trace of the matrix $\textbf{A}$. The mean trace represents the average probability that a randomly selected annotator provides an accurate label. Intuitively, minimizing the trace encourages the estimated annotators to be maximally unreliable while minimizing the cross entropy ensures fidelity with observed noisy annotators. We minimize this combined loss via stochastic gradient descent to learn both $\{\theta, \phi\}$. 

\noindent\textbf{Learning with Negative Coarse Label.} For some situations, it is easier to provide the negative coarse annotation, e.g., complementary label, to help ML model predict the true label distribution. Thus, we study a readily available substitute, namely complementary labeling. However, if we still use traditional loss functions $\mathcal{L}_{\text{obj}}(\theta, \phi)$ when learning with these complementary labels, similar to Eq.~(\ref{eq4}), we can only learn a mapping $\mathbb{R} \rightarrow \mathcal{Y}$ that tries to predict conditional probabilities $p(\tilde{\textbf{y}}^{(c)} \mid \textbf{x})$ and the corresponding complementary pixels classifier that predicts a $\tilde{y}_{wh}^{(c)}$ for a given observation $\mathbf{x}$. 

To address the above issue, inspired by Yu \cite{yu2018learning}, which summarizes all the probabilities into a transition matrix $\mathbf{M} \in \mathbb{R}^{L \times L}$, where $\textbf{m}(\textbf{x},w,h)_{i j}:=p(\tilde{y}^{\{c\}}_{wh}=i\mid y_{wh}=j,\textbf{x})$ and $\textbf{m}(\textbf{x},w,h)_{i i}=0, \forall i, j \in \{1,...,L\}$. Here, $\textbf{m}_{i j}$ denotes the entry value in the $i$-th row and $j$-th column of $\mathbf{M}$. As shown in Fig.~\ref{pipeline}, we achieve this simply by adding a linear layer to the complementary label learning channel. This layer outputs $v(\textbf{x})$ by multiplying the output of the CE function (i.e., $u(\textbf{x})$ ) with the transposed transition matrix $\mathbf{M}^{\top}$. Note that the transition matrix is also widely exploited in Markov chains \cite{gagniuc2017markov} and has many applications in machine learning, such as learning with label noise \cite{tanno2019learning,zhang2020disentangling,zhang2020learning}.

Recall that in transition matrix $\mathbf{M}, \textbf{m}_{i j}=p(\tilde{y}^{\{c\}}_{wh}=i\mid y_{wh}=j,\textbf{x})$ and $\textbf{m}_{i i}=p(\tilde{y}^{\{c\}}_{wh}=i\mid y_{wh}=i,\textbf{x})=0$. We observe that $p(\tilde{\textbf{y}}^{\{c\}} \mid \textbf{x})$ can be transferred to $p(\tilde{\textbf{y}}^{\{c\}} \mid \textbf{x})$ by using the transition matrix $\mathbf{M}$,
$$
\begin{aligned}
p(\tilde{y}_{wh}^{(c)}=j \mid \textbf{x}) &=\sum_{i \neq j} p(\tilde{y}_{wh}^{(c)}=j, \bar{y}_{wh}^{(c)}=i \mid \textbf{x}) \\
&=\sum_{i \neq j} p(\tilde{y}_{wh}^{(c)}=j \mid \bar{y}_{wh}^{(c)}=i, \textbf{x}) p(\bar{y}_{wh}^{(c)}=i \mid \textbf{x}) \\
&=\sum_{i \neq j} p(\tilde{y}_{wh}^{(c)}=j \mid \bar{y}_{wh}^{(c)}=i) p(\bar{y}_{wh}^{(c)}=i \mid \textbf{x})
\end{aligned}
$$
Intuitively, if $\textbf{v}_i(\textbf{x})$ tries to predict the probability $p(\tilde{y}^{(c)}=i \mid \textbf{x}), \forall i \in[1,...,L]$, then $\mathbf{M}^{-\top} \textbf{v}$ can predict the probability $p(\tilde{y}^{(o)} \mid \textbf{x})$, which is the positive prediction of the corresponding complementary coarse label. To enable end-to-end learning rather than transferring after training, we let
$$
\mathbf{v}(\textbf{x})=\mathbf{M}^{\top} \mathbf{u}(\textbf{x})
$$
where $\mathbf{u}(\textbf{x})$ is now an intermediate output of the complementary coarse annotation, and $\mathcal{L}_{\text{comp}}(\theta, \phi)=\arg \max _{i \in[L]} \textbf{v}_i(\textbf{x})$. Then, the modified loss function $\bar{\mathcal{L}}_{\text{obj}}(\theta, \phi)$ is
\begin{equation}
\begin{aligned}
&\bar{\mathcal{L}}_{\text{obj}}(\theta, \phi)(\textbf{u}(\textbf{x}), \tilde{y}^{(c)}) := {\mathcal{L}}_{\text{comp}}(\theta, \phi)(\mathbf{v}(\textbf{x}), \hspace{1mm}\tilde{y}^{(c)})\\
&:= {\mathcal{L}}_{\text{comp}}(\theta, \phi)(\mathbf{M}^{\top}\cdot \{\hat{\textbf{A}}_{\phi}^{(c)}(\textbf{x}_n)\cdot \hat{\textbf{p}}_\theta ({\textbf{x}_n})\}, \hspace{1mm} \tilde{y}^{(c)})  \\ 
    &:= \sum\limits_{n = 1}^N {\sum\limits_{c = 1}^C \mathds{1}(c \in \mathcal{S}({\textbf{x}_n}))} \cdot \Big{[} \text{CE}\big{(}\mathbf{M}^{\top}\cdot \{\hat{\textbf{A}}_{\phi}^{(c)}(\textbf{x}_n)\cdot \hat{\textbf{p}}_\theta({\textbf{x}_n})\},\hspace{1mm} \tilde{\textbf{y}}_n^{(c)}\big{)}\hspace{1mm} \\
    & + \hspace{1mm}\lambda\cdot \text{tr}\big{(}\hat{\textbf{A}}^{(c)}_{\phi}(\textbf{x}_n)\big{)}\Big{]}
\end{aligned}
\end{equation}
In this way, if we can learn an optimal $\mathbf{v}$ such that $\textbf{v}_i(\textbf{x})=p(\tilde{y}^{(c)}=i \mid \textbf{x}), \forall i \in[L]$, meanwhile, we can also find the optimal $\mathbf{u}$ and the loss function ${\mathcal{L}}_{\text{comp}}(\theta, \phi)$, which can be easily applied to deep learning. With sufficient training examples with complementary coarse labels, this DNN often simultaneously learns good classifiers for both $(\textbf{x}, \tilde{y}^{(c)})$ and $(\textbf{x}, \tilde{y}^{(o)})$.

Finally, we combine the positive annotation loss ${\mathcal{L}}_{\text{obj}}$ and the negative annotation loss ${\mathcal{L}}_{\text{comp}}$ as our objective and optimize:
\begin{equation}
{\mathcal{L}}_{\text{final}}(\theta, \phi):= {\mathcal{L}}_{\text{obj}}(\theta, \phi) + {\mathcal{L}}_{\text{comp}}(\theta, \phi).
\end{equation}

\section{Experiments}

In this section, we first describe our dataset and then show the coarse annotation refinement scenarios in simulated and real-world settings, separately.

\textbf{Datasets.} MNIST dataset \cite{lecun1998gradient} consists of 60,000 training and 10,000 testing examples, all of which are 28 $\times$ 28 grayscale images of digits from 0 to 9, and we derive the segmentation labels by thresholding the intensity values at 0.5. The Cityscapes \cite{cordts2016cityscapes} dataset contains 5000 high-resolution (2048 $\times$ 1024 pixels) urban scene images collected across 27 European Cities. The dataset comprises 5,000 fine annotations (2,975 for training, 500 for validation, and 1,525 for testing) and 20,000 coarse annotations where 11,900 samples for training and 2,000 for validation (i.e., coarse polygons covering individual objects). The LES-AV \cite{orlando2018towards} is a dataset for retinal vessel segmentation on color fundus images. It comprises 22 fundus photographs with available manual annotations of the retinal vessels including annotations of arteries and veins. The 22 images/patients are acquired with resolutions of 30$^\circ$ field-of-view (FOV) and 1444 $\times$ 1620 pixels (21 images), and 45$^\circ$ FOV and 1958 $\times$ 2196 pixels (one image), with each pixel $= 6\mu m$. We divide them into 18 images for training and 4 images for testing.

\textbf{Synthetic Noisy Coarse Annotations.} 
We generate synthetic coarse noisy annotations from an assumed expert consensus label on MNIST, Cityscapes and retinal fundus image dataset, to demonstrate the efficacy of the approach in an idealized situation where the expert consensus label is known. We simulate the positive and negative coarse noisy annotations by performing morphological transformations (e.g., thinning, thickening, fractures, etc) on the expert consensus label and background (complementary label), using Morpho-MNIST software \cite{castro2019morphomnist}. In particular, \textit{positive coarse noisy annotation} is prone to be poor segmentation, which is simulated by combining small fractures and over-segmentation; \textit{negative coarse noisy annotation} always annotates on the background or complementary label using the same approach. 
To create synthetic coarse noisy labels in the multi-class scenario, we use a similar simulation to create coarse labels on the Cityscapes dataset. We first choose a target class and then apply morphological operations on the provided coarse mask to create the two synthetic coarse labels at different patterns, namely, objective coarse and complementary coarse annotations. We create training data by deriving labels from the simulated annotations.

\begin{figure*}[t]
    \centering
    \includegraphics[width=\linewidth]{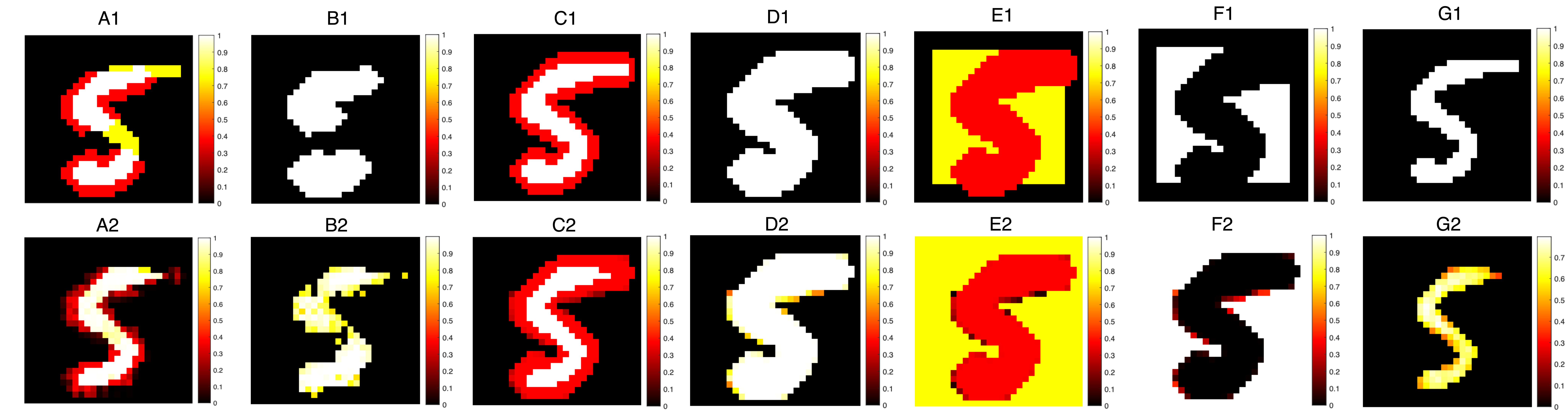}
    \caption{Visualisation of the estimated labels, the estimated pixel-wise CMs, and the estimated TMs on MNIST datasets along with their targets (best viewed in color). White is the true positive, yellow is the false negative, red is the false positive, and black is the true negative. A1-A2: the target and estimated CM: $\textbf{A}^{(o)}_\phi(\textbf{x})$ for positive coarse annotation; B1-B2: the given and estimated positive coarse annotation $\tilde{\textbf{y}}^{(o)}$; C1-C2: the target and estimated intermediate CM: $\textbf{A}^{(c)}_\phi(\textbf{x})$ for negative coarse annotation; D1-D2: the target and estimated intermediate negative coarse annotation \textbf{u}(\textbf{x}); E1-E2: the target and estimated TM: $\mathbf{M}^{\top}$ for negative coarse annotation; F1-F2: the provided and estimated negative coarse annotation ($\tilde{\textbf{y}}^{(c)}$ and $\textbf{v}(\textbf{x})$); G1-G2: target label and our estimation.}
    \label{MNIST-CM}
\end{figure*}

\textbf{Comparison Methods and Evaluation Metrics.}
Our experiments are based on the assumption that no expert consensus label is available a priori, hence, we compare our method against multiple weakly-supervised and semi-supervised methods. In particular, we explore our method with ablation studies, e.g., our method without negative coarse annotation; we also consider two popular interactive image segmentation algorithms for generating masks from scribbles: GrabCut \cite{rother2004grabcut} and LazySnapping \cite{li2004lazy}, then training FCNs using the masks generated by these methods. Meanwhile, we compare our weakly-supervised results based on the noisy coarse annotations and strongly-supervised results based on the expert consensus annotations. For evaluation metrics, we use mIoU between estimated segmentation $\hat{\textbf{p}}_\theta(\textbf{x})$ and expert consensus label ${\textbf{y}}_{GT}$.

\textbf{Strategy performance of utilizing coarse annotation.} Our method jointly propagates information into unmarked pixels and learns network parameters. An easy way is to first use any existing interactive image segmentation methods to generate masks based on coarse annotation, and then use these masks to train FCNs. In Table~\ref{scribbleresults}, we compare our methods with these two-step solutions. We investigate two popular interactive image segmentation algorithms for generating masks from coarse annotation: GrabCut \cite{rother2004grabcut} and LazySnapping \cite{li2004lazy}. Given the coarse annotations, GrabCut generates the mask only for the target pixels while LazySnapping produces the masks not only for the objective but also for the non-target pixels. Training FCNs using the masks generated by these methods shows inferior semantic segmentation accuracy. This is because these traditional methods \cite{rother2004grabcut,li2004lazy} only focus on the low-level color/spatial information and are unaware of semantic content. The generated masks cannot be the reliable ``GT'' for training the supervised networks. On the contrary, our coarse-based supervised method achieves a score of 82.5\% on MNIST and 68.3\% on Cityscapes dataset, about 10\% higher than the two-step solutions. This is because our model can capture the patterns of mistakes for each noisy coarse annotation, and the high-level information can help with the coarse-to-fine propagation of the true label estimation. This behavior is shown in Fig.~\ref{MNIST-CM}.

\begin{table}[t]
\centering
%\scriptsize
%\setlength{\tabcolsep}{3.1mm}
%\renewcommand{\arraystretch}{1.5}
\begin{tabular}{c|c|c}
\toprule
 Methods    & MNIST &  Cityscapes  \\  \midrule
GrabCut + FCN   & 75.2 $\pm$ 0.3  & 53.6 $\pm$ 0.4 \\  
LazySnapping+FCN  & 78.5 $\pm$ 0.2  & 59.4 $\pm$ 0.4 \\  \midrule
Ours (w/o negative annotation)    & 77.2 $\pm$ 0.2  & 62.3 $\pm$ 0.2 \\ 
Ours (Full)    &  82.5 $\pm$ 0.1  & 68.3 $\pm$ 0.2 \\ \bottomrule
\end{tabular}
\caption{Segmentation results (mIoU (\%)) on the MNIST and Cityscapes validation set via different strategies of utilizing coarse annotations.}\label{scribbleresults}
\end{table}

\begin{figure*}[t]
    \centering
    \includegraphics[width=\linewidth]{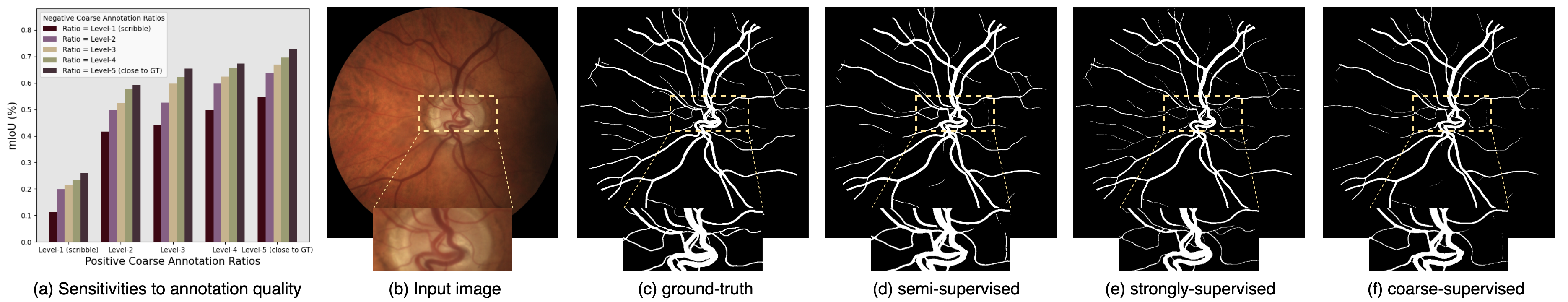}
    \caption{Sensitivities to quality of positive and negative coarse annotations, and the visualization of the estimated labels with different supervision approaches.}
    \label{retinal_results}
\end{figure*}

\textbf{Performance on Retinal Vessel Segmentation.}
We illustrate the results of our approach on a more challenging dataset with real coarse and noisy labels from the medical domain. This dataset, called LES-AV, consists of images of the retinal fundus acquired from different patients. The task is to segment the vessel into a binary mask (see Fig.~\ref{retinal_results}). The process of segmenting the blood vessel in the retinal image is crucial for the early detection of eye diseases.

An experienced annotator was tasked with providing the practical positive and negative coarse annotations for each sample on LES-AV dataset. We generate such a real-world dataset to show the segmentation results and evaluate the performance of different supervision approaches. Meanwhile, we also created 5 different ratio levels for the positive and negative coarse annotations from \textit{level-1} (tend to scribble) to \textit{level-5} (tend to GT) with increasing ratios compared to the given expert consensus labels. We use such a dataset to evaluate the sensitivity to annotation quality of our model on medical image data.

We show the results of sensitivities to annotation quality in Fig.~\ref{retinal_results}(a). 
Our model performs robustly and gradually improved when the ratio of coarse annotations is increased. Especially when the ratio is increased from \textit{level-1} to \textit{level-2}, our model's performance is increased significantly and comparable to the mask-level results. By applying our practical annotations, we conduct a group of experiments under different supervision. The results in Table.~\ref{retinal_supervision} indicate that our WSL approach achieves comparable results to the strongly-supervised method. Meanwhile, by including some extra coarse annotations, the result is improved 3\%. Finally, we present the segmentation visualization in Fig.~\ref{retinal_results}(d$\sim$f).

\begin{table}[t]
\centering
%\scriptsize
%\setlength{\tabcolsep}{3.1mm}
%\renewcommand{\arraystretch}{1.5}
\begin{tabular}{l|ll|c|c} 
\toprule
supervision &  w/ masks &  w/ coarse & total & mIoU (\%) \\
\midrule weakly & $-$ & $16$ & $16$ & 65.8 $\pm$ 0.3 \\
strongly & $16$ & $-$ & $16$ & 69.2 $\pm$ 0.2\\
semi & $16$ & $4$  & $20$ & 71.6 $\pm$ 0.3\\
\bottomrule
\end{tabular}
\caption{Comparisons of our method using different annotations on the LES-AV retinal image validation set. The term ``w/ masks'' shows the number of training images with mask-level annotations, and ``w/ coarse'' shows the number of training images with coarse annotations.}\label{retinal_supervision}
\end{table}

\section{Conclusion}

We introduced a new theoretically grounded algorithm for recovering the expert consensus label distribution from noisy coarse annotations. Our method enjoys implementation simplicity, requiring only adding a complementary label learning term to the loss function. Experiments on both synthetic and real data sets have shown superior performance over the common WSL and SSL methods in terms of both segmentation accuracy and robustness to the quality of coarse annotations and label noise. Furthermore, the method is capable of estimating coarse annotations even when scribble is given per image.

% Our work was primarily motivated by medical imaging applications for which pixel-level annotation is required for a large number of images. These include segmentation of retinal vessels on fundus photography, organs on computed tomography, and cells on histopathology slides. However, future work shall consider imposing structures on the CMs and TMs to broaden the applicability to scribble or spot annotations, which could contribute to saving the labor of the labeling process. Another limiting assumption is to learn only from the coarse annotation in difficult cases. The majority of segmentation failures happen in difficult cases or within difficult portions of images. Only providing coarse annotations for these difficult cases or portions and learning from them is also a valuable next step.

Our work was primarily motivated by medical imaging applications that require pixel-level annotation for a large number of images. These include segmentation of retinal vessels on fundus photography, organs on computed tomography, and cells on histopathology slides. However, future work shall consider imposing structures on the CMs and TMs to broaden the applicability to scribble or spot annotations, which could contribute to saving the labor of the labeling process. Another promising direction is to explore the use of promptable foundation models \cite{simons2024spinefm} for image segmentation, which have recently shown strong potential in handling a variety of input forms and tasks. Incorporating such models into our framework could further enhance flexibility and reduce annotation burdens.

\section*{Acknowledgment}
This work was supported by Novo Nordisk.

\begin{comment}  %% removed for anonymized MICCAI 2025 submission.
    
    % The following acknowledgement and disclaimer sections should be removed for the double-blind review process.  
    % If and when your paper is accepted, reinsert the acknowledgement and the disclaimer clause in your final camera-ready version.

\begin{credits}
\subsubsection{\ackname} A bold run-in heading in small font size at the end of the paper is
used for general acknowledgments, for example: This study was funded
by X (grant number Y).

\subsubsection{\discintname}
It is now necessary to declare any competing interests or to specifically
state that the authors have no competing interests. Please place the
statement with a bold run-in heading in small font size beneath the
(optional) acknowledgments\footnote{If EquinOCS, our proceedings submission
system, is used, then the disclaimer can be provided directly in the system.},
for example: The authors have no competing interests to declare that are
relevant to the content of this article. Or: Author A has received research
grants from Company W. Author B has received a speaker honorarium from
Company X and owns stock in Company Y. Author C is a member of committee Z.
\end{credits}

\end{comment}

\bibliographystyle{splncs04}  
\bibliography{references}

\end{document}